# RRV: A Spatiotemporal Descriptor for Rigid Body Motion Recognition

Yao Guo, Youfu Li, *Senior Member, IEEE*, and Zhanpeng Shao

*Abstract*—The motion behaviors of a rigid body can be characterized by a six degrees of freedom motion trajectory, which contains the 3-D position vectors of a reference point on the rigid body and 3-D rotations of this rigid body over time. This paper devises a rotation and relative velocity (RRV) descriptor by exploring the local translational and rotational invariants of rigid body motion trajectories, which is insensitive to noise, invariant to rigid transformation and scale. The RRV descriptor is then applied to characterize motions of a human body skeleton modeled as articulated interconnections of multiple rigid bodies. To show the descriptive ability of our RRV descriptor, we explore its potentials and applications in different rigid body motion recognition tasks. The experimental results on benchmark datasets demonstrate that our RRV descriptor learning discriminative motion patterns can achieve superior results for various recognition tasks.

## I. Introduction

TO ACHIEVE effective robot–human interaction, robots need to be able to recognize the motion behaviors of objects or individuals from their visual observations. Motion trajectories produced by objects and individuals provide abundant clues on mining spatiotemporal information in motion characterization [1]. In recent years, a number of 3-D trajectory-based gait recognition [2], 3-D handwritten recognition [3], gesture recognition [4]–[6], and human action recognition [7], [8] approaches have been proposed. One of the inherent challenges in trajectory-based recognition is the various variations of the raw data induced by the viewpoint changing, noise contamination or being performed by different individuals. An invariant descriptor can offer substantial advantages over the raw data in capturing spatiotemporal features [5], [6]. Most of the previous descriptors are mainly proposed for 3-D point trajectories matching and recognition [5], [6], [9]. However, motion behaviors of an object cannot be uniquely characterized by the 3-D motion trajectory of a reference point on the object, which is insufficient for description without considering rotations of this object.

In this paper, the 6-D rigid body motion trajectory, which can be characterized by a set of 3-D position vectors of a reference point on the rigid body and the 3-D rotations of this rigid body over time [10], will be taken into account. The description of 6-D rigid body motion trajectories has attracted increased attentions in recent years, where the instantaneous screw axis (ISA) descriptor [11] and the SoSaLe descriptor [12] have achieved some successful applications. The ISA descriptor is a 6-D representation based on a motion model of the ISA. Two of six invariants are the translational and rotational speeds along the ISA, and the other four invariants involve the first-order and second-order kinematics of the ISA. As for the SoSaLe descriptor, the position frame and the orientation frame attached to a rigid body are built via Frenet–Serret formulas [13]. Two invariants in SoSaLe descriptor are the translational and rotational speeds, and the other four are angle variations of two frames between two adjacent time instances. However, these two descriptors still have some limitations. They both involve second-order time derivatives of discrete trajectories, which is sensitive to noise. Moreover, the applications on multiple rigid bodies are not investigated.

To address the above challenges, this paper first devises a rotation and relative velocity (RRV) descriptor for representing 6-D rigid body motion trajectories. Our RRV descriptor, capturing the local spatiotemporal features, only involves first-order time derivatives of the trajectory. Specifically, the RRV descriptor consists of 4-D rotational invariants and 3-D translational invariants. The rotational invariant part is the reparameterized unit quaternion for describing the 3-D rotations of rigid bodies. Meanwhile, the square-root velocity function (SRVF) [14], [15] of the 3-D point trajectory is the translational invariants by being projected onto the instantaneous rotation matrix, which can further explore the internal dynamics between the position vectors and 3-D rotations. A flexible metric is then introduced to measure the distance between two RRV descriptors.

In recent years, human action recognition from the 3-D data has attracted widespread attentions [7], [8], [16]–[24]. The representations of the 3-D data can be separated into two groups: 1) depth image [16]–[19] and 2) 3-D skeleton [7], [8], [20]–[25]. Depth images provide sufficient

Manuscript received November 3, 2016; accepted May 5, 2017. This work was supported in part by the Research Grant Council of Hong Kong under Grant CityU11205015, and in part by the Natural Science Foundation of China under Grant 61673329 and Grant 61603341. This paper was recommended by Associate Editor B. W. Schuller. *(Corresponding author: Youfu Li.)*
Y. Guo and Y. Li are with the Department of Mechanical and Biomedical Engineering, City University of Hong Kong, Hong Kong (e-mail: yaoguo4-c@my.cityu.edu.hk; meyfli@cityu.edu.hk).
Z. Shao is with the College of Computer Science and Technology, Zhejiang University of Technology, Hangzhou 310023, China (e-mail: zpshao@zjut.edu.cn).
Color versions of one or more of the figures in this paper are available online at http://ieeexplore.ieee.org.
Digital Object Identifier 10.1109/TCYB.2017.2705227

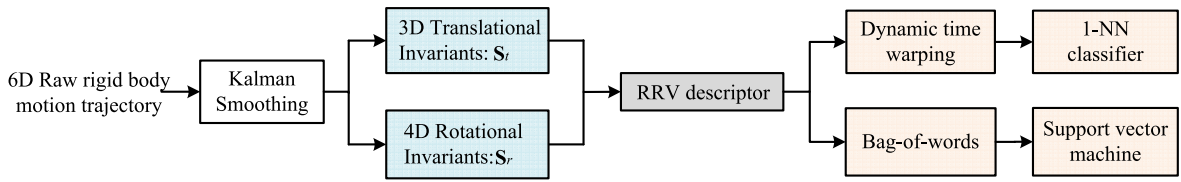

Fig. 1. Pipeline of rigid body motion recognition approaches using the RRV descriptor. A 6-D raw trajectory will be smoothed by Kalman smoother to remove noise and outliers. Following this, the RRV descriptor for representing this trajectory will be separated into two parts: translational invariants $\mathbf{S}_t$ and rotational invariants $\mathbf{S}_r$. For recognition, we use DTW approach to measure the similarity between two RRV sequences and the 1-NN classifier is chosen to recognize the test samples. We also introduce a computational efficient BoW based recognition method in this paper.

geometric features of pixels which capture the point clouds of the human body and the background in 3-D space [18]. The 3-D skeleton of a human body captured by RGB-D sensors [26] or motion capture system [27] also have been intensively studied in human action representations due to the robustness to variations of viewpoint, human body scale and motion speed as well as the real-time performance [20], [21]. In this paper, we extend the RRV descriptor to multiple rigid bodies for skeleton-based human action recognition. We first define a human body coordinate system, where the joint trajectories can be represented in this local coordinate system. As the skeleton of a human body can be modeled as articulated interconnections of multiple rigid bodies, we decompose the whole skeleton into five hierarchical parts in representing different actions. Specifically, instead of the direct interconnections of rigid bodies in each part, a novel virtual rigid body (VRB) is introduced to provide the discriminative representation for each part, which can improve the discriminability in the description. The RRV descriptors for representing different hierarchical body parts are concatenated together as the final skeleton representations for action recognition.

For various rigid body motion recognition tasks, the dynamic time warping (DTW) approach measures the similarity between two RRV sequences and the nearest neighbor (1-NN) classifier is applied. To improve the computational efficiency, we also encode RRV sequences with Bag-of-Words (BoW) [28] approach as the input of the support vector machine (SVM) classifier. Finally, our method is evaluated on the benchmark AUSLAN2 [29], MSRAction3D [16], and MSRC-12 datasets [30] to demonstrate the descriptive power of our RRV descriptor for both single rigid body motion recognition and skeleton-based human action recognition tasks. Fig. 1 is the pipeline of the proposed approach. In the conclusion, the main contributions of this paper are threefold as follows.

1) A new invariant RRV descriptor with noise robustness is proposed to represent 6-D rigid body motion trajectories.
2) The RRV descriptor is extended to characterize motions of a human body skeleton modeled as articulated interconnections of multiple rigid bodies. To our knowledge, this is the first application of descriptors incorporating rotational invariants on the multiple rigid bodies' motion recognition.
3) The proposed VRB provide a discriminative and compact representation for characterizing the motions of each hierarchical part of a skeleton.

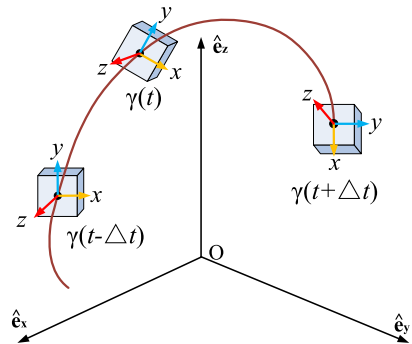

Fig. 2. 6-D representation of a rigid body motion trajectory. It can be characterized by the position vectors of a reference point on the rigid body and 3-D rotations of this rigid body over time.

This paper is organized as follows. In Section II, we derive the RRV descriptor by exploring local rotational and translational invariants of 6-D rigid body motion trajectories. In Section III, we discuss the properties of our RRV descriptor. A flexible metric is also introduced to measure the distance between two RRV descriptors. The RRV descriptor is extended to represent the motion behaviors of a human body's skeleton in Section IV. The recognition approaches incorporating the RRV descriptor are addressed in Section V. The experimental studies on benchmark datasets are conducted to demonstrate the effectiveness and consistency of the RRV descriptor in Section VI.

## II. RRV Descriptor

### A. Rigid Body Motion Trajectory

A rigid body is a solid body in which the deformation is neglected. Most of the moving objects can be viewed as rigid bodies. Understanding rigid body motion trajectories can provide sufficient clues on mining spatiotemporal information in motion characterization. The motion behaviors of a rigid body can be characterized by a six degrees of freedom (DOF) trajectory [10]. A 6-D rigid body motion trajectory $\mathbf{m}(t)$ parameterized by time index $t$ in Fig. 2 can be written as

$$\mathbf{m}(t) = \begin{bmatrix} \boldsymbol{\gamma}(t) \\ \boldsymbol{\Theta}(t) \end{bmatrix}, t \in [1, N] \qquad (1)$$

where $\boldsymbol{\gamma}(t) = [x(t), y(t), z(t)]^{\mathrm{T}}$ is the position vector of a reference point on the rigid body recorded in the world coordinate system {E}, and $\boldsymbol{\Theta}(t)$ represents the 3-D rotations of this rigid body across time. Various representations exist to express a rotation in 3-D space. The popular ones are in the form of Euler angles, rotation matrix, and unit quaternion. These representations still have three DOF, even though they use more than the necessary minimum of three parameters.

This paper aims to propose an invariant and gradient-based descriptor $\mathbf{S}$ by capturing the local rotational and translational invariants of rigid body motion trajectories, which can be invariant to undesirable variations over the raw data and offer sufficient descriptive power. It should be noted that the rotations and position vectors are measured in different scales. Therefore, this new descriptor $\mathbf{S}$ is divided into the following two parts: 1) rotational invariant part $\mathbf{S}_r$ and 2) translational invariant part $\mathbf{S}_t$.

### B. Rotational Invariants in RRV Descriptor

According to Euler's rotation theorem [31], any rotation in 3-D space is equivalent to a rotation by an angle $\beta$ with respect to a fixed unit Euler axis $\hat{\mathbf{w}} = [\hat{w}_1, \hat{w}_2, \hat{w}_3]^T$. $\hat{\mathbf{w}}$ is a unit vector denoting the direction of rotation, and $\beta$ denotes the magnitude of rotation along the axis $\hat{\mathbf{w}}$. Then, the axis-angle vector can be defined as

$$\mathbf{w} = \beta\hat{\mathbf{w}} = [\beta\hat{w}_1, \beta\hat{w}_2, \beta\hat{w}_3]^T \tag{2}$$

which is also called rotation vector or Euler vector [31]. In mathematics, this rotation vector $\mathbf{w}$ parameterizes a 3-D rotation in 3-D Euclidean space. Moreover, such representation has the minimum of three real parameters.

Considering the vision capture system is rotated, then the rotated coordinate system is $\{\boldsymbol{\Gamma}E\}$, where $\boldsymbol{\Gamma}$ indicates a random rotation matrix. Mathematically, $\boldsymbol{\Gamma}$ belongs to the special orthogonal group SO($n$), where we have $\boldsymbol{\Gamma}^T\boldsymbol{\Gamma} = \boldsymbol{\Gamma}\boldsymbol{\Gamma}^T = \mathbf{I}$ and $\det(\boldsymbol{\Gamma}) = 1$. Regarding the change of perspectives or the rotational variations of trajectories while being performed by different individuals, the angle $\beta$ remains a constant value. However, the direction of rotation $\hat{\mathbf{w}}$ will vary in $\{\boldsymbol{\Gamma}E\}$, which means that $\boldsymbol{\Theta}(t)$ is not invariant to rotational variations. To this end, the singular value decomposition (SVD)-based rotational normalization step will be first adopted.

*1) Rotational Normalization:* Denote a matrix $\mathbf{A} = [\hat{\mathbf{w}}(t_1), \hat{\mathbf{w}}(t_2), \ldots, \hat{\mathbf{w}}(t_N)]$ as the combination of the Euler axis $\hat{\mathbf{w}}$ for all time instances, where $\mathbf{A} \in \mathbb{R}^{3\times N}$ is the field of real numbers. By taking SVD, $\mathbf{A}$ can be decomposed as

$$\text{svd}(\mathbf{A}) = \mathbf{U\Sigma V}^T \tag{3}$$

where $\mathbf{U}$ is a $3 \times 3$ orthogonal matrix. $\boldsymbol{\Sigma}$ has the same dimensions as $\mathbf{A}$, which is a $3 \times N$ rectangular matrix with non-negative real numbers on the diagonal. $\mathbf{V}^T$ is the transpose of an $N \times N$ matrix $\mathbf{V}$. As is well known, the columns of $\mathbf{V}$ are the eigenvectors of $\mathbf{A}^T\mathbf{A}$. The entries on the diagonal of $\boldsymbol{\Sigma}$ are singular values, which are the square roots of the eigenvalues of $\mathbf{A}^T\mathbf{A}$. In addition, we have

$$(\boldsymbol{\Gamma}\mathbf{A})^T(\boldsymbol{\Gamma}\mathbf{A}) = \mathbf{A}^T(\boldsymbol{\Gamma}^T\boldsymbol{\Gamma})\mathbf{A} = \mathbf{A}^T\mathbf{A}. \tag{4}$$

It can be observed that $\mathbf{A}^T\mathbf{A}$ is invariant to rotational variations. In other words, $\mathbf{A}^T\mathbf{A}$ and $(\boldsymbol{\Gamma}\mathbf{A})^T(\boldsymbol{\Gamma}\mathbf{A})$ share the same eigenvectors and eigenvalues. Consequently, the matrices $\boldsymbol{\Sigma}$ and $\mathbf{V}^T$ remain unchanged under rotational variations.

In this paper, $\mathbf{U}$ can be viewed as a reparameterized rotation matrix. Then the rotational variations of $\mathbf{A}$ can be removed by prerotating it with $\mathbf{U}^T$ as

$$\tilde{\mathbf{A}} = \mathbf{U}^T\mathbf{A} = \boldsymbol{\Sigma}\mathbf{V}^T = [\tilde{\mathbf{w}}(t_1), \tilde{\mathbf{w}}(t_2), \ldots, \tilde{\mathbf{w}}(t_N)] \tag{5}$$

where $\tilde{\mathbf{w}} = \mathbf{U}^T\hat{\mathbf{w}}$ is the reparameterization of the unit Euler axis $\hat{\mathbf{w}}$.

*2) Rotation Invariants $\mathbf{S}_r$:* Given the reparameterized unit vector $\tilde{\mathbf{w}}$ and angle $\beta$, an appropriate representation for 3-D rotations can be obtained accordingly. It is important to note that Euler angles and rotation matrix have some limitations in constructing a flexible and compact descriptor. With Euler angles, the 3-D rotation needs to be realized in a specified order. Moreover, a rotation matrix uses more than the necessary minimum of parameters. On the contrary, unit quaternion offers a convenient representation for 3-D rotations [32]. Therefore, the rotational invariants $\mathbf{S}_r$ in our RRV descriptor is the reparameterized unit quaternion calculated by $\tilde{\mathbf{w}}$ and $\beta$, which can be written as

$$\mathbf{S}_r = \tilde{\mathbf{q}} = [q_w, q_x, q_y, q_z]^T = \left[\cos\frac{\beta}{2}, \tilde{\mathbf{w}}^T\sin\frac{\beta}{2}\right]^T \tag{6}$$

where $\mathbf{S}_r$ is a 4-D representation vector at each time step.

### C. Translational Invariants in RRV Descriptor

The 3-D point trajectory is a set of position vectors of a reference point on the rigid body, which can be parameterized by the time index $t$

$$\boldsymbol{\gamma}(t) = [x(t), y(t), z(t)]^T, \ t \in [1, N]. \tag{7}$$

Similarly, raw trajectories will suffer rigid transformation and scale in 3-D space while being captured from different perspectives. Moreover, the variability in shape and the motion speed usually happen when different objects or individuals perform the same trajectory. Hence, the translational invariant part $\mathbf{S}_t$ investigates the local invariant spatiotemporal features of 3-D point trajectories can offer substantial advantages over raw data.

*1) Shape Analysis of 3-D Curves:* The SRVF [14], [15] has been proposed for representing the shapes of 3-D curves and achieves some successful applications in trajectory matching and recognition. SRVF is a gradient-based representation, which has shown discriminative ability in the description. The length of 3-D trajectories will be normalized as one to remove scale influences. Then the SRVF of a 3-D trajectory $\boldsymbol{\gamma}(u)$ is given as

$$\boldsymbol{\beta}(u) = \frac{\dot{\boldsymbol{\gamma}}(u)}{\sqrt{\|\dot{\boldsymbol{\gamma}}(u)\|}} \tag{8}$$

where $u$ indicates an arbitrary parameter, $\|\cdot\|$ is $l_2$-norm and $\dot{\boldsymbol{\gamma}}(u)$ denotes the first-order derivative with respect to $u$.

It is clear that the original trajectories, as well as the corresponding SRVFs, are not invariant to rotational variation. Motivated by this, the translational invariant part $\mathbf{S}_t$ in the RRV descriptor is expected to extend the standard SRVF to its rotational invariant version. Compare to removing rotation variations in a dynamic programming method [15] with high computational cost, we can calculate the SRVF of 3-D point trajectory in a local coordinate system.

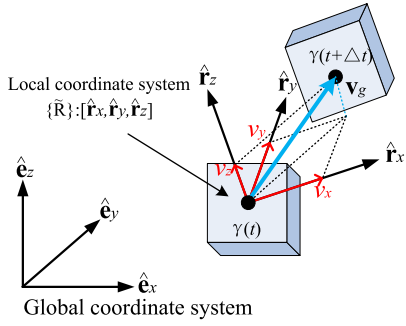

Fig. 3. Velocity vector of the point trajectory will be projected onto the three axes of the view-independent local coordinate system $\{\tilde{\mathbf{R}}\}$.

*2) View-Independent Local Coordinate System:* Commonly, the rigid body coordinate system is a fixed one attached to the rigid body, which is predefined by users. However, the rigid body coordinate system may cannot be consecutively extracted from all time instances due to occlusions. Some works denote the Frenet–Serret frame [6], [12] as the local coordinate system, which can be built according to the first-order and second-order time derivatives of the motion trajectory. Although it can be acquired across all time instances, the Frenet–Serret frame is sensitive to noise, speed variations, and local perturbations of the trajectory.

In this paper, we denote the instantaneous rotation matrix $\mathbf{R}(t)$ at time $t$ as a view-independent local coordinate system. This local coordinate system is not only the intrinsic feature of the motion trajectory but also robust to noise and speed variations. As discussed above, the reparameterized unit quaternion $\tilde{\mathbf{q}}$ introduced in (6) is invariant to rotational variations. Then the reparameterized rotation matrix $\tilde{\mathbf{R}} = [\hat{\mathbf{r}}_x, \hat{\mathbf{r}}_y, \hat{\mathbf{r}}_z]$ is a view-independent representation and can be derived from $\tilde{\mathbf{q}}$

$$\tilde{\mathbf{R}} = \begin{bmatrix} 1 - 2q_y^2 - 2q_z^2 & 2q_x q_y - 2q_z q_w & 2q_x q_z + 2q_y q_w \\ 2q_x q_y + 2q_z q_w & 1 - 2q_x^2 - 2q_z^2 & 2q_y q_z - 2q_x q_w \\ 2q_x q_z - 2q_y q_w & 2q_y q_z + 2q_x q_w & 1 - 2q_x^2 - 2q_y^2 \end{bmatrix}.$$

Denote $_{\{\Gamma E\}}\mathbf{v}_g = [v_x, v_y, v_z]^T$ be the velocity of the 3-D point trajectory $\tilde{\boldsymbol{\gamma}}(t)$ represented in the rotated coordinate system $\{\Gamma E\}$, where $\tilde{\boldsymbol{\gamma}}(t) = \mathbf{U}^T \boldsymbol{\gamma}(t)$ and $\mathbf{U}^T$ is calculated by (3). As shown in Fig. 3, $_{\{\Gamma E\}}\mathbf{v}_g$ is then projected onto three principal axes $\hat{\mathbf{r}}_x$, $\hat{\mathbf{r}}_y$, and $\hat{\mathbf{r}}_z$, which can be expressed as

$$_{\{\tilde{R}\}}\mathbf{v}_l = \tilde{\mathbf{R}}^T \mathbf{v}_g = \left[\hat{\mathbf{r}}_x^T \mathbf{v}_g, \hat{\mathbf{r}}_y^T \mathbf{v}_g, \hat{\mathbf{r}}_z^T \mathbf{v}_g\right]^T \quad (9)$$

where $_{\{\tilde{R}\}}\mathbf{v}_l$ is the velocity vector expressed in the coordinate system $\{\tilde{\mathbf{R}}\}$. This simple projection suffices to capture the internal dynamics between rotational and translational features of the 6-D rigid body motion trajectories, thus to improving the descriptiveness power. Finally, the translational invariants $\mathbf{S}_t$ in the RRV descriptor at each time point is the SRVF of the local velocity vector $_{\{\tilde{R}\}}\mathbf{v}_l$, which is

$$\mathbf{S}_t = \frac{\mathbf{v}_l}{\sqrt{\|\mathbf{v}_l\|}} \quad (10)$$

where $\mathbf{S}_t$ is a 3-D representation vector.

*D. RRV Descriptor*

In the conclusion, this paper devises a flexible and invariant descriptor for representing 6-D rigid body motion trajectories, which is called RRV descriptor. The RRV descriptor at each time step has two parts for representing the rotational and translational invariants

$$\mathbf{S} = \left[\mathbf{S}_r^T, \mathbf{S}_t^T\right]^T \quad (11)$$

where $\mathbf{S}_r \in \mathbb{R}^{4 \times 1}$ is the rotational invariant part in the form of the reparameterized unit quaternion, and $\mathbf{S}_t \in \mathbb{R}^{3 \times 1}$ is the translational invariant part that calculates the SRVF of the point trajectory in a local coordinate system.

## III. INVARIANTS AND PROPERTIES

### A. Invariant Properties

As we claim, the RRV descriptor can show strong invariance to various transformations in 3-D space. In the following, the invariant properties of the RRV descriptor will be declared.

*1) Invariant to Translation:* The velocity vectors and 3-D rotations of a rigid body will remain unchanged in terms of the translation in 3-D space. Hence, the RRV descriptor $\mathbf{S}$ is invariant to translation.

*2) Invariant to Scale:* The scale can be simplified as the zoom in and zoom out of 3-D point trajectories in visual observations, where 3-D rotations are not influenced by scale. In our method, the point trajectories will be normalized as unit length to remove scale.

*3) Invariant to Rotation:* Rotational variations of rigid body motion trajectories are mainly attributed to different viewpoints of vision systems. As presented in Section II, the matrix $\mathbf{A} = [\hat{\mathbf{w}}(t_1), \hat{\mathbf{w}}(t_2), \ldots, \hat{\mathbf{w}}(t_N)]$ represents a sequence of directions of 3-D rotations. Then we have svd$(\mathbf{A}) = \mathbf{U\Sigma V}^T$ by applying SVD. Hence, the rotational variations of $\hat{\mathbf{w}}$ can be removed by rotating it with $\mathbf{U}^T$ as $\tilde{\mathbf{w}}$. The rotational invariants $\mathbf{S}_r$ is the unit quaternion derived by the $\tilde{\mathbf{w}}$ and $\beta$. For $\mathbf{S}_t$, the SRVF of $\tilde{\boldsymbol{\gamma}}(t)$ is represented in a view-independent local coordinate system $\{\tilde{R}\}$, which is also invariant to rotation.

To demonstrate the invariant properties of our RRV descriptor, the RRV descriptors of two rigid body motion trajectories are given in Fig. 4. To compare with the trajectory $\mathbf{m}_1$ (blue cross), the trajectory $\mathbf{m}_2$ (red circle) is scaled, rotated, and translated and the white Gaussian noise is added to $\mathbf{m}_1$. As can be observed, the corresponding RRV descriptors $\mathbf{S}_1$ and $\mathbf{S}_2$ match well with each other under various variations.

### B. Metric Between RRV Descriptors

An appropriate metric for calculating the distance between two RRV descriptors contributes to improving the matching and recognition performance. The previous works [11], [12] calculate the $l_2$-norm between two descriptors, which is insufficient due to rotational and translational invariants are measured in different scales. In this paper, we introduce a flexible metric for calculating the distance between two RRV descriptors. Given two RRV descriptors $\mathbf{S}^p$ and $\mathbf{S}^q$, the

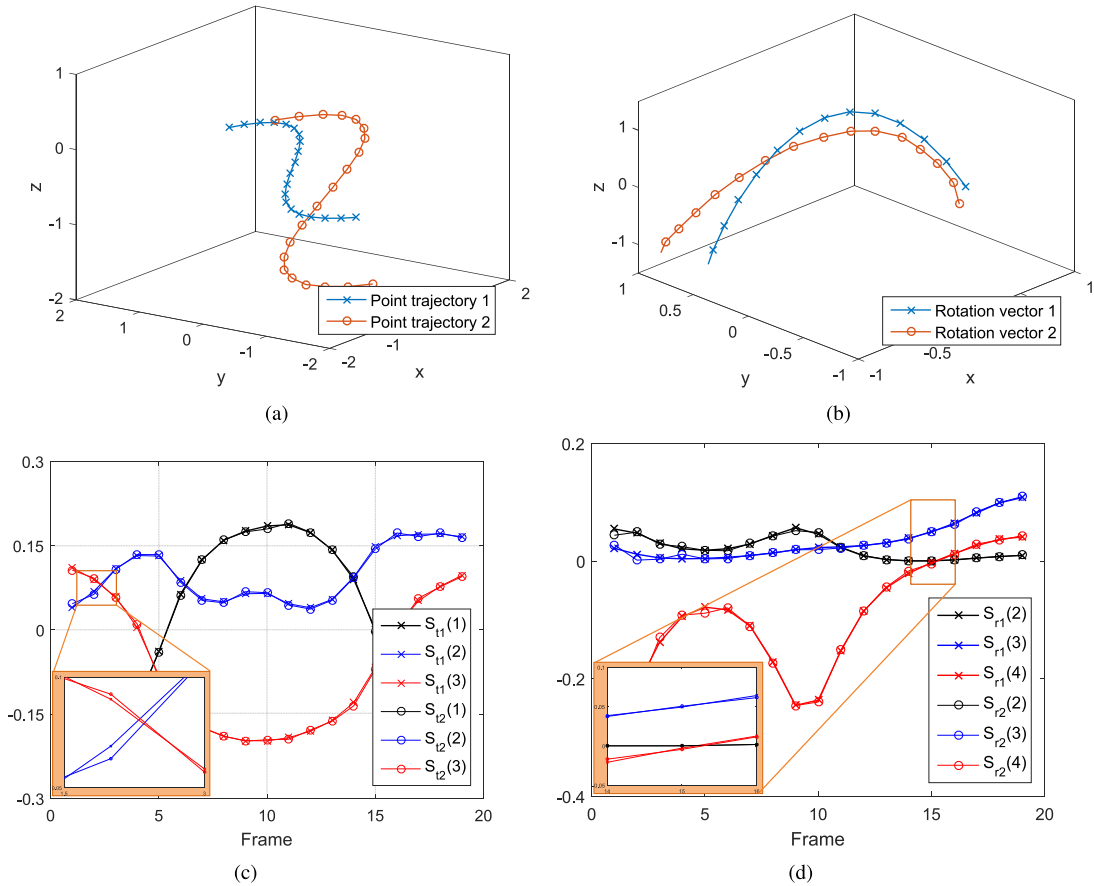

Fig. 4. Two rigid body motion trajectories $\mathbf{m}_1$ and $\mathbf{m}_2$, and the corresponding RRV descriptors $\mathbf{S}_1$ and $\mathbf{S}_2$. Compare with the trajectory $\mathbf{m}_1$ (blue cross line), the $\mathbf{m}_2$ (red circle line) is scaled, rotated, translated, and additive white Gaussian noise is added to it. (a) 3-D point trajectories. (b) Two sequences of rotation vectors indicating 3-D rotations. (c) Two translational invariants $\mathbf{S}_t$. (d) Corresponding rotational invariants $\mathbf{S}_r$ of two trajectories. The angle $\beta$ remains unchanged under various variations, so we only demonstrate the last three values in each $\mathbf{S}_r$.

distance $d(\mathbf{S}^p, \mathbf{S}^q)$ is defined by separating the rotational and translational invariants

$$d(\mathbf{S}^p, \mathbf{S}^q) = \min\{\|\mathbf{S}_r^p - \mathbf{S}_r^q\|, \|\mathbf{S}_r^p + \mathbf{S}_r^q\|\} + \|\mathbf{S}_t^p - \mathbf{S}_t^q\| \quad (12)$$

where the first part denotes the distance between two unit quaternions [33], and the second part indicates the distance between two local SRVFs.

### C. Special Patterns

Next, we discuss some special patterns in RRV descriptor.

*1) Rotation With Incomplete Observations:* In some situations, the rigid body is a cylinder or a conic with similar textures. Instead, 3-D rotations of this rigid body can be calculated by the movement of a principal axis $\hat{\mathbf{b}}_x$, where rotations along $\hat{\mathbf{b}}_x$ will not be taken into consideration. Given $\hat{\mathbf{b}}_x(t)$ and $\hat{\mathbf{b}}_x(t+1)$, we can denote $\hat{\mathbf{a}}_1 = \hat{\mathbf{b}}_x(t)/\|\hat{\mathbf{b}}_x(t)\|$ and $\hat{\mathbf{a}}_2 = \hat{\mathbf{b}}_x(t+1)/\|\hat{\mathbf{b}}_x(t+1)\|$ as two unit vectors. The cross-product of these two vectors is $\mathbf{c} = \hat{\mathbf{a}}_1 \times \hat{\mathbf{a}}_2$. Then the rotation matrix $\mathbf{R}(t)$ can be calculated by

$$\mathbf{R}(t) = \mathbf{I} + [\mathbf{c}]_\times + [\mathbf{c}]_\times^2 \frac{1 - \hat{\mathbf{a}}_1 \cdot \hat{\mathbf{a}}_2}{\|\mathbf{c}\|_2^2} \quad (13)$$

where $[\mathbf{c}]_\times$ denotes the skew-symmetric cross product of $\mathbf{c} = [c_1, c_2, c_3]^T$

$$[\mathbf{c}]_\times = \begin{bmatrix} 0 & -c_3 & c_2 \\ c_3 & 0 & -c_1 \\ -c_2 & c_1 & 0 \end{bmatrix}. \quad (14)$$

*2) Pure Rotation:* The position vector $\boldsymbol{\gamma}(t)$ at time step $t$ is equal to the previous one $\boldsymbol{\gamma}(t-1)$ when a pure rotation acts on, then the translational velocity $\mathbf{v}_g(t) = 0$ for all time instances. Accordingly, the translational invariants $\mathbf{S}_t$ are equal to $[\mathbf{0}, \mathbf{0}, \mathbf{0}]^T$ of length $N$, where $\mathbf{0}$ represents the column vector with all zero elements.

*3) Pure Translation:* Under pure translation, the rotational invariants $\mathbf{S}_r$ will be $[\mathbf{1}, \mathbf{0}, \mathbf{0}, \mathbf{0}]^T$, where $\mathbf{1}$ and $\mathbf{0}$ are the column vectors of length $N$ with the entire entries equal to one and zero, respectively. Moreover, the local coordinate system can be regarded as the identity matrix.

*4) Predefined Local Coordinate System:* A fixed local coordinate system can be defined according to the geometric feature of the rigid body or the external information. For instance, the motion trajectory of a hand viewed as a rigid body can be represented in a predefined human body coordinate system. Along this line, the initial rigid body motion trajectory is already invariant to rotation, which suggests that the SVD-based rotational normalization step and the projection of SRVF can be skipped in this special case.

## IV. Extensions on Multiple Rigid Bodies

The mechanism of applying the RRV descriptor for representing motion trajectories of multiple rigid bodies will be investigated in this paper. First, considering a simple scenario that $L$ rigid bodies are free of connection, the integrated descriptor concatenate $\mathbf{S}_i$ ($i = 1, 2, \ldots, L$) for each rigid body as $\mathbf{S}_{\text{Multi}} = [\mathbf{S}_1^T, \mathbf{S}_2^T, \ldots, \mathbf{S}_L^T]^T$.

In this paper, we mainly focus on the more challenging articulated interconnections of multiple rigid bodies, which can be widely applied in modeling the skeleton of human bodies. Human action recognition from 3-D skeleton data is one of the indispensable topics in human–robot interactions [20], [21]. The skeleton data, containing the 3-D position vectors of a set of key joints in each frame, can be extracted by some low-cost RGB-D sensors [26] (Kinect, Realsense, etc.) or motion capture system [27]. On the other hand, some works can achieve similar action recognition from depth images [19], [34] capturing the point clouds of the human body and background in 3-D space. It should be noted that this paper on RRV descriptor focuses on skeleton-based action recognition so as to compare our approach with the previous skeleton-based works [8], [22]–[25].

### A. Skeleton of Human Body

The skeleton of a human body captured by Kinect v1 consists of the 3-D position vectors of 20 key joints $J_n = [x_n, y_n, z_n]^T$ ($n = 1, 2, \ldots, 20$), as demonstrated in Fig. 5. Without of loss generality, each interconnection between two adjacent joints can be assumed as a rigid body. Let $\mathbf{b}_x$ be the vector indicating the direction from one joint to another joint, and it locates on the surface of this rigid body.

Given a skeleton, we can define a view-independent human body coordinate system $\{H\} = [\hat{\mathbf{h}}_x, \hat{\mathbf{h}}_y, \hat{\mathbf{h}}_z]$ using the joint positions so as to represent motion trajectories in this coordinate system. Taking the skeleton in Fig. 5 as the example, the hip center joint $J_7$ is set as the origin of the human body coordinate system as shown in Fig. 6. Denote the vector between shoulder center $J_3$ and right hip $J_5$ as $\mathbf{v}_1$ and the vector between $J_3$ and left hip $J_6$ as $\mathbf{v}_2$, respectively. Then the human body coordinate system $\{H\}$ can be derived by

$$\begin{cases} \mathbf{h}_y = \mathbf{v}_1 + \mathbf{v}_2, & \hat{\mathbf{h}}_y = \dfrac{\mathbf{h}_y}{\|\mathbf{h}_y\|} \\ \mathbf{h}_z = \mathbf{v}_1 \times \mathbf{v}_2, & \hat{\mathbf{h}}_z = \dfrac{\mathbf{h}_z}{\|\mathbf{h}_z\|} \\ \mathbf{h}_x = \mathbf{h}_y \times \mathbf{h}_z, & \hat{\mathbf{h}}_x = \dfrac{\mathbf{h}_x}{\|\mathbf{h}_x\|} \end{cases} \quad (15)$$

where $\mathbf{h}_y$ is the sum of $\mathbf{v}_1$ and $\mathbf{v}_2$ pointing up to the shoulder, $\mathbf{h}_z$ referring to the face direction is the normal vector of the plane determined by the joints $\{J_3, J_5, J_6\}$, and $\mathbf{h}_x$ is the cross product of $\mathbf{h}_y$ and $\mathbf{h}_z$. It is clear that the human body coordinate system depends on the human body itself. The joint trajectories expressed in $\{H\}$ are invariant to rotation, which means that the rotational normalization step in calculating the RRV descriptor can be skipped.

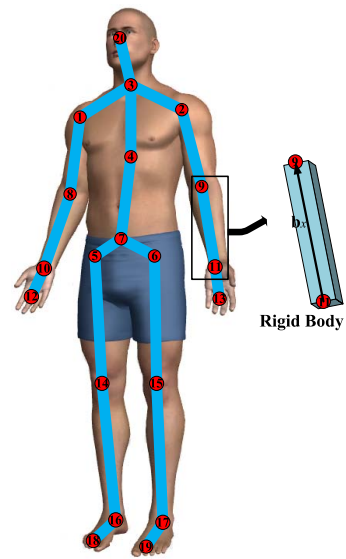

Fig. 5. Example of the human body's skeleton extracted by Kinect v1. The articulated interconnection between two adjacent joints can be assumed as a rigid body. Then the whole skeleton is modeled as articulated interconnections of multiple rigid bodies.

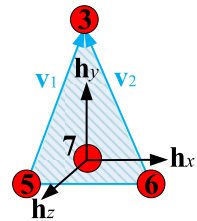

Fig. 6. Human body coordinate system $\{H\}$ of a skeleton. The hip center $J_7$ is set as the origin.

It is important to note that human actions typically involve different parts of a human body [35]. To effectively integrate our RRV descriptor for multiple rigid bodies, a hierarchical skeleton could be helpful to produce levels of meaningful parts for representing different actions. Moreover, the hierarchical decomposition is beneficial for dealing with the situations that the execution rates of each part are distinct in the same action and the symmetric parts are used to perform the same action.

### B. Hierarchical Structure of the Skeleton

As illustrated in Fig. 7, the skeleton of a human body can be decomposed into five hierarchical parts as: 1) left arm (LA); 2) right arm (RA); 3) torso (TS); 4) left leg (LL); and 5) right leg (RL). In other words, the action of a human body is the combination of the motions of these five parts. Especially, the movement of the TS is represented by the 6-D rigid body motion trajectory of the human body coordinate system $\{H\}$ in this paper. Without loss of generality, we can assume that rigid bodies belong to the same part are correlated to each other while performing a certain action, which means that the orientations and positions of these rigid bodies will change simultaneously.

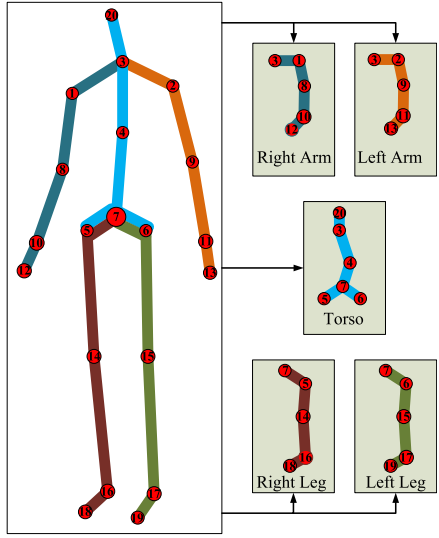

Fig. 7. Skeleton of a human body can be decomposed into five body parts: LA, RA, TS, LL, and RL. The motion behavior of the TS can be represented by the 6-D rigid body motion trajectory of the human body coordinate system $\{H\}$.

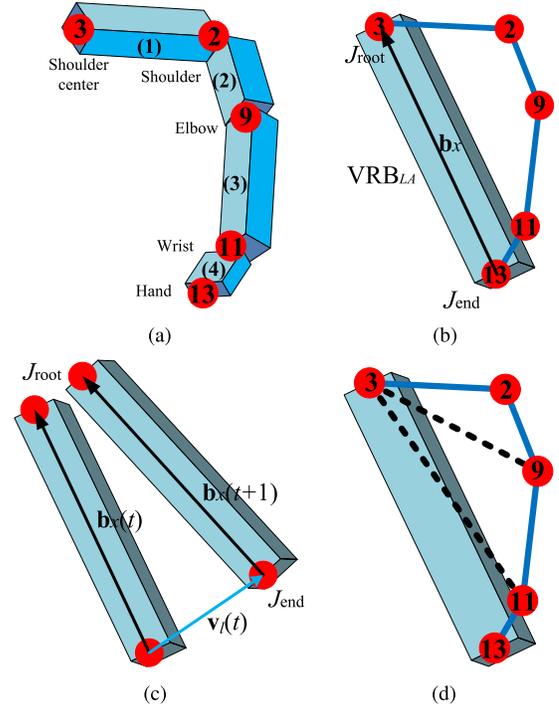

Fig. 8. Demonstration of constructing a VRB for representing the LA. (a) Simple representation for an LA are articulated interconnections of four multiple rigid bodies. (b) VRB can be constructed with respect to the root and end joints. The end point is chosen as the reference point. (c) Illustration of 3-D rotation of this VRB and the velocity vector of the reference point. (d) Other candidate virtual rigid bodies (black dashed line) in the hierarchical component LA.

Fig. 8(a) demonstrates an example of an LA consists of four rigid bodies. Commonly, the integrated descriptor for each part can directly concatenate descriptors of $L$ rigid bodies together as the final feature. Then the concatenated RRV descriptor for representing the LA can be written as $\mathbf{S}_{LA} = [\mathbf{S}_1^T, \mathbf{S}_2^T, \mathbf{S}_3^T, \mathbf{S}_4^T]^T$ and $\mathbf{S}_{LA} \in \mathbb{R}^{28 \times N}$. However, the simple concatenation will produce ambiguity in some special situations and lead to unreasonable recognition results. Besides, this representation also has high dimensions, thus to increasing the computational cost.

### C. Virtual Rigid Body

To improve the descriptive power and reduce dimensions in characterizing multiple rigid bodies in each part, we introduce a new VRB approach in this paper. Instead of the reality rigid bodies as shown in Fig. 8(a), a VRB is constructed to provide an alternative representation of each hierarchical part with multiple rigid bodies. As indicated in Fig. 8(b), we can denote a root joint $J_{\text{root}}$ and an end joint $J_{\text{end}}$ in each part, respectively. Without loss of generality, these two joints can be assumed to locate on the surface of this VRB, and the point $J_{\text{end}}$ is chosen as the reference point.

It is important to note that the VRB proposed here is just an analogical concept, where the distance between $J_{\text{root}}$ and $J_{\text{end}}$ will not be limited to a constant value. Denote the vector indicating the direction from $J_{\text{root}}$ to $J_{\text{end}}$ as

$$\mathbf{b}_x = J_{\text{root}} - J_{\text{end}}. \quad (16)$$

According to the definition of the RRV descriptor, the key step is to estimate the 3-D rotations of this VRB and calculate the velocity vectors of $J_{\text{end}}$. However, the other two principal axes $\hat{\mathbf{b}}_y$ and $\hat{\mathbf{b}}_z$ cannot be uniquely determined only given two points on this rigid body. Hence, the 3-D rotations of this rigid body can be calculated by $\hat{\mathbf{b}}_x = \mathbf{b}_x / \|\mathbf{b}_x\|$ across two adjacent time instances as discussed in Section III-C. Given a pair of $\mathbf{b}_x(t)$ and $\mathbf{b}_x(t+1)$ at time instant $t$ and $t+1$ as illustrated in Fig. 8(c), instantaneous rotation matrix $\mathbf{R}(t)$ can be calculated according to (13). As introduced above, the whole skeleton is invariant to rotation while representing the joint trajectories in the local coordinate system $\{H\}$, which suggests that the SVD-based rotational normalization step can be skipped in calculating the RRV descriptor for each VRB. Then the rotational invariants $\mathbf{S}_r$ in the RRV descriptor can be derived with the given $\mathbf{R}(t)$

$$\begin{cases} \mathbf{S}_r = \mathbf{q} = [q_w, q_x, q_y, q_z]^T \\ q_w = \frac{1}{2}\sqrt{1 + \text{tr}(\mathbf{R})}, \quad q_y = \frac{1}{4q_w}(\mathbf{R}_{13} - \mathbf{R}_{31}) \\ q_x = \frac{1}{4q_w}(\mathbf{R}_{32} - \mathbf{R}_{23}), \quad q_z = \frac{1}{4q_w}(\mathbf{R}_{21} - \mathbf{R}_{12}). \end{cases} \quad (17)$$

Next, the velocity vector $\mathbf{v}_l(t)$ can be determined with the adjacent position vectors of the end joint as shown in Fig. 8(c). Accordingly, the translational invariants $\mathbf{S}_t$ in the RRV descriptor is derived by (8).

By introducing this VRB method, the latent spatiotemporal correlations of the joints in each component can be captured, which improve the richness in the description. Similarly, other candidate virtual rigid bodies (black dashed lines) in $LA$ also can be constructed as listed in Fig. 8(d), where the root joint is fixed. However, it is important to balance the tradeoff between the descriptive power and the compactness of the final descriptor for representing human actions. As illustrated in [36] and [37], those acral joints, denoting as the informative joints, play more important role in skeleton-based action recognition tasks. Hence, the joints located at the extremity of the skeleton are preferred in constructing the virtual rigid bodies in this paper.

### D. RRV-Based Skeleton Representation

Similarly, the descriptors $\mathbf{S}_{RA}$, $\mathbf{S}_{LL}$, and $\mathbf{S}_{RL}$ for the RA, LL, and RL can be built, and $\mathbf{S}_{TS}$ is the RRV descriptor for representing the rigid body motion trajectory of the local coordinate system $\{H\}$. In summary, the final representation for characterizing the skeleton at each time instant is the concatenation of the descriptors of five components, which is $\mathbf{S}_{\text{human}} = [\mathbf{S}_{LA}^T, \mathbf{S}_{RA}^T, \mathbf{S}_{TS}^T, \mathbf{S}_{LL}^T, \mathbf{S}_{RL}^T]^T$.

### E. Metric Between Multiple RRV Descriptors

Considering the descriptor $\mathbf{S}_{\text{Multi}} = [\mathbf{S}_1^T, \mathbf{S}_2^T, \ldots, \mathbf{S}_L^T]^T$ containing $L$ RRV descriptors, the distance between two integrated descriptors $\mathbf{S}_{\text{Multi}}^p$ and $\mathbf{S}_{\text{Multi}}^q$ is the summation of $d(\mathbf{S}_i^p, \mathbf{S}_i^q)$ introduced in (12) for each pair $\mathbf{S}_i^p$ and $\mathbf{S}_i^q$ ($i = 1, 2, \ldots, L$). It can be concluded as

$$d(\mathbf{S}_{\text{Multi}}^p, \mathbf{S}_{\text{Multi}}^q) = \sum_{i=1}^{L} d_i(\mathbf{S}_i^p, \mathbf{S}_i^q). \tag{18}$$

## V. RIGID BODY MOTION RECOGNITION

### A. Dynamic Time Warping-Based Recognition

As the rigid body motion trajectories are often parameterized by time index $t$, our RRV descriptor is also generated as a time sequence. To compare the similarity between two temporal sequences of unequal size, DTW is a popular method. Based on the idea of dynamic programming, the sequences will be nonlinearly warped to explore the minimum cost path in the distance matrix $\mathbf{D}$, which can be written as

$$\begin{aligned}\mathbf{D}(p, q) = &\ d(p, q) \\ &+ \min\{\mathbf{D}(p-1, q), \mathbf{D}(p, q-1), \mathbf{D}(p-1, q-1)\}\end{aligned} \tag{19}$$

where $d(p, q)$ is the metric introduced in (12). The minimum accumulative cost of aligning two sequences will be recorded in the bottom right entry of $\mathbf{D}$. In the applications of skeleton-based recognition, the DTW method will be separately applied for warping RRV sequence in the each part to find a minimum cost path. This is beneficial for dealing with the execution rates for different parts are distinct while different individuals perform the same action. Moreover, we also calculate the similarity between two actions with the same parts and with the symmetric parts to determine the final warping cost.

For matching and recognition, a testing sequence will be labeled as the same class as its 1-NN in terms of the minimum cost among all the training sequences.

### B. Bag-of-Words-Based Recognition

In this paper, a BoW-based recognition method is also proposed to balance the tradeoff between the computational cost and recognition accuracy. It is clear that RRV descriptor sequences are of unequal size. Recently, BoW [28] approach becomes popular since it can provide a feasible technique to match samples with unequal sizes and interclass variations. According to BoW approach, a single feature vector of length $m$ can be called as a patch. For all given patches from the whole train data, the $K$ cluster centroids can be offline learned by a fast k-means clustering method [38]. These $K$ vectors will be composed of a code book or a dictionary $[\mathbf{W}_i]_{i=1:K} \in \mathbb{R}^{K \times m}$, where each vector in the dictionary can be denoted as a "visual word." The fundamental idea of BoW is to find the existence of each visual word in a given test sample. Hence, given a testing patch, the standard $l_2$-norm from the patch to the $K$ cluster centroid $[\mathbf{W}_i]_{i=1:K}$ will be calculated, thus to finding the nearest neighbor. Finally, all the patches in a sample will be encoded as a sparse histogram $\mathbf{H} \in \mathbb{R}^{K \times 1}$ as the input to the classifier, in which the $i$th bin of $\mathbf{H}$ represents the occurrence frequency of the word $\mathbf{W}_i$. However, temporal information will be neglected during the transformation from time sequences to histogram, which leads to a loss in recognition accuracy.

Recall that the RRV descriptor $\mathbf{S}$ for each time instant has the rotational invariants $\mathbf{S}_r \in \mathbb{R}^{4 \times 1}$ and the translational invariants $\mathbf{S}_t \in \mathbb{R}^{3 \times 1}$. Motivated by this, two dictionaries $[\mathbf{W}_i]_{i=1:K_r}^r$ and $[\mathbf{W}_j]_{j=1:K_t}^t$ for these two parts will be established separately, where the sizes of the two dictionaries are $K_r$ and $K_t$, respectively. As a result, two sparse histograms $\mathbf{H}^r \in \mathbb{R}^{K_r \times 1}$ and $\mathbf{H}^t \in \mathbb{R}^{K_t \times 1}$ will be concatenated as $\mathbf{H} = [\mathbf{H}^r, \mathbf{H}^t]$, and $\mathbf{H} \in \mathbb{R}^{(K_r+K_t) \times 1}$ is the final representation for each RRV descriptor sequence. As for multiple RRV descriptors representing the skeleton-based action, two sparse histograms $\mathbf{H}^r$ and $\mathbf{H}^t$ regarding rotational and translational invariants are constructed for five hierarchical parts, respectively. Then the histograms for the five body parts will be concatenated as the final representation for the whole skeleton.

For multiple class classification, we train the nonlinear SVM using the chi-square kernel, which is beneficial for classifying the histogram features. Finally, recognition results will be given based on a one-against-one rule.

## VI. EXPERIMENTS

To demonstrate the effectiveness of our RRV descriptor, we conduct both single and multiple rigid bodies motion recognition experiments on different benchmark datasets so as to compare our approach with the previous methods.

### A. Experimental Setup

*1) Australian Sign Language (AUSLAN2) Dataset:* The meaning of a sign language can be given by the motions of two hands and ten fingers. Moreover, the relations between two hands or other body parts also provide informative clues in the representation [39]. It should be noted that we focus on evaluating the effectiveness of our RRV descriptor for representing single rigid body motion trajectory on AUSLAN2 dataset so as to compare with previous trajectory descriptors. To this end, each palm can be viewed as a single rigid body, and the relations between two hands are not taken into consideration in this paper. The 6-D motion trajectory of each palm, including the position vectors $\boldsymbol{\gamma}(t) = [x(t), y(t), z(t)]^T$ of a point in the middle of wrist hand and Euler angles

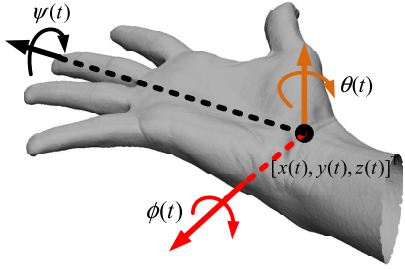

Fig. 9. 6-D motion trajectory of the right palm, where the reference point is the middle point of the wrist.

TABLE I
ROOT AND END JOINTS OF VIRTUAL RIGID BODIES IN RRV-BASED SKELETON REPRESENTATION

|  | Left arm | | Right arm | | Left leg | | Right leg | |
|---|---|---|---|---|---|---|---|---|
| $J_{root}$ | 3 | 3 | 3 | 3 | 7 | 7 | 7 | 7 |
| $J_{end}$ | 11 | 13 | 10 | 12 | 19 | 17 | 18 | 16 |

$\Theta(t) = [\phi(t), \psi(t), \theta(t)]^T$, are recorded in the AUSLAN2 dataset [29] as shown in Fig. 9.

The AUSLAN2 dataset has 95 classes, where nine subjects (three samples in each subject) for each class are performed by different individuals to improve intraclass variation. To show the comparative results with the ISA [11] and SoSaLe [12] descriptors, we also implement them on AUSLAN2 dataset. Similarly, the 1-NN classifier with DTW method is adopted for classifying ISA and SoSaLe sequences. The integrated descriptor for each sign language can be written as $\mathbf{S}_{\text{sign}} = [\mathbf{S}_{\text{LH}}^T, \mathbf{S}_{\text{RH}}^T]^T$, where the subscripts "LH" and "RH" are short for the left hand and right hand, respectively. To demonstrate the robustness and consistency of our approach, the twofold cross-subject (CS) experiments of 95 classes are conducted, where four subjects are selected as train data and the remaining five subjects are chosen for testing.

The following three setups using RRV descriptor will be evaluated on AUSLAN2 dataset.
1) *RRV-$l_2$+DTW:* The standard $l_2$-norm is to measure the distance between two RRV descriptors.
2) *RRV-New+DTW:* The distance between two RRV descriptors is measured by the proposed metric as in (12). The setups 1) and 2) aim to demonstrate the effectiveness of the new metric. The 1-NN classifier with DTW method calculating the similarity between RRV sequences is applied.
3) *RRV+BoW:* BoW encodes the RRV sequences of different sizes into the histograms of the same length as the input to the SVM classifier.

The sizes of the two dictionaries are set as $K_r = 120$ and $K_t = 130$ in a heuristic way.

*2) MSRAction3D Dataset:* The MSRAction3D dataset is a popular dataset for human action recognition, where many researchers have reported their results on this dataset using skeleton data, depth images, or the combination. It should be noted that our approach focuses on utilizing RRV descriptors for skeleton-based human action recognition [20], [21] rather than the fusion of depth images. Hence, we compare our RRV-based skeleton representation with other state-of-the-art skeleton-based action recognition approaches.

The MSRAction3D dataset consists of twenty actions: 1) *high arm wave*; 2) *horizontal arm wave*; 3) *hammer*; 4) *hand catch*; 5) *forward punch*; 6) *high throw*; 7) *draw X*; 8) *draw tick*; 9) *draw circle*; 10) *hand clap*; 11) *two hand wave*; 12) *side boxing*; 13) *bend*; 14) *forward kick*; 15) *side kick*; 16) *jogging*; 17) *tennis swing*; 18) *tennis serve*; 19) *golf swing*; and 20) *pick up and throw*. Each action has ten subjects (2–3 samples in each subject) performed by different individuals. The 3-D positions of 20 joints are extracted from the consecutive depth images by using the real-time skeleton tracking algorithm [40]. The total number of available samples is 557, where the remaining 43 sequences have missing or wrong skeletal data [22]. We follow the more challenging cross subject setting as mentioned in [16] and [22]. The subjects 1, 3, 5, 7, and 9 are chosen for training and subjects 2, 4, 6, 8, and 10 are for testing. To illustrate the robustness of our approach, the cross-validation with all possible 5-5 splits (252 in total) are also performed.

In the conclusion, the following four setups with RRV descriptor for skeleton-based action recognition will be evaluated on MSRAction3D dataset. The distance between two RRV descriptors is calculated by the metric as given in (12).
1) *RRV+LCS+DTW:* In this setup, the final representation of each body part is the concatenation of the RRV descriptors of the reality rigid bodies belonging to this part as shown in Fig. 8(a).
2) *RRV-VRB+LCS+DTW:* The discriminative VRB method is utilized for representing each body part. The setups 1) and 2) aims to evaluate the descriptive power of VRB method. LCS is short for the local coordinate system, which means that all the joint trajectories are represented in the human body system $\{H\}$ defined in Section IV-A. The SVD-based rotational normalization and the projection of SRVF are skipped in setups 1) and 2). As suggested in Table I, the following eight virtual rigid bodies and the human body coordinate system $\{H\}$ are selected for representing the five hierarchical parts.
3) *RRV-VRB+GCS+DTW:* GCS stands for the global coordinate system, which means that the SVD-based rotational normalization step is applied for each VRB motion trajectory. The comparison of setup 2) and 3) is to demonstrate the effectiveness of the skeleton normalization method using the human body coordinate system.
4) *RRV-VRB+LCS+BoW:* The BoW method encodes the RRV sequences of the five hierarchical parts into a histogram of the same length. This histogram is the input to the SVM classifier. The sizes of the two dictionaries are $K_r = 120$ and $K_t = 180$, respectively.

*3) MSRC-12 Dataset:* To test our proposed approach when a large number of action instances are available, we conduct the experiments on the MSRC-12 dataset [30]. MSRC-12 dataset is a relatively large dataset for 3-D skeleton-based action recognition. This data has 594 sequences, containing the performance of 12 action classes by 30 subjects. One action is performed several times by the same individual in each sequence. To realize the action recognition experiments, we follow the annotation of the sequences in [7] to mark the onset and offset of action instance in each sequence. Table II lists 12 action classes in the dataset and the number of annotated action instances of each class.

TABLE II
ACTIONS AND NUMBER OF INSTANCES IN MSRC-12 DATASET

| Metaphoric gestures | No. of insts. | Iconic gestures | No. of insts. |
|---|---|---|---|
| Start system | 508 | Duck | 500 |
| Push right | 522 | Goggles | 508 |
| Wind it up | 649 | Shoot | 511 |
| Bow | 507 | Throw | 515 |
| Had enough | 508 | Change weapon | 498 |
| Beat both | 516 | Kick | 502 |

TABLE III
COMPARISON RESULTS OF THE RRV DESCRIPTOR WITH EXISTING DESCRIPTORS ON AUSLAN2 DATASET

| Methods | Mean±STD(%) |
|---|---|
| SoSaLe+DTW | 73.27±2.43 |
| ISA+DTW | 82.94±2.64 |
| RRV-$l_2$+DTW | 87.45±2.09 |
| RRV+BoW | 90.92±2.61 |
| **RRV-new+DTW** | **92.56±2.04** |

TABLE IV
COMPUTATIONAL COST OF DIFFERENT APPROACHES ON AUSLAN2 DATASET

| Methods | Time cost per sample (ms) |
|---|---|
| RRV-new+DTW | 417 |
| ISA+DTW | 398 |
| SoSaLe+DTW | 406 |
| **RRV+BoW** | **24** |

First of all, we follow the leave-one-subject-out (LOSubO) protocol as given in [7] and [41]. Following this, all actions belong to 29 subjects are chosen for training, and the remaining one subject is used for testing, which means that the recognition experiments will be repeated 30 times. The benefit of this setup is that our approach can be evaluated using as much training data as possible. Moreover, it helps to analyze the source subject of the classification errors. Next, we also conduct the experiments with a more challenging CS protocol as introduced in [7] and [25], which can evaluate the sensitivity of the approach with reduced training samples. In this setup, the odd subjects are for training and the action instances in even subjects are chosen as testing data.

In summary, we evaluate the following two methods using RRV descriptor for skeleton-based action recognition on MSRC-12 dataset. The distance between two RRV descriptors is measured by (12).

1) *RRV-VRB+LCS+DTW:* The VRB is utilized for providing the discriminative representative for each hierarchical component. All the joint trajectories are represented in the coordinate system {H}. Because the skeleton in MSRC-12 is also captured by Kinect v1, the same virtual rigid bodies are constructed as listed in Table I.
2) *RRV-VRB+LCS+BoW:* The size of dictionaries with respect to rotational and translational invariants are set as $K_r = 180$ and $K_t = 180$, respectively.

4) *Implementation:* Raw trajectories are usually contaminated with noise and need to be smoothed before devising the description. The smoothing process may change the shape of input trajectories more or less. To balance the tradeoff between smoothing effects and shape preservation, it is necessary to choose an appropriate smoother. Kalman smoother is a suitable filter which can remain the essential shape of a trajectory as well as smooth the noisy points. In addition, some raw trajectories may record a fraction of stationary sequences at the beginning or end, and these fractions make no contribution in motion characterization but increase the computational cost. Inspired by this, the stationary data will be removed in the preprocessing step. All the experiments in this paper are implemented in MATLAB R2013a and on a desktop with Intel(R) Corel(TM) i5-2400 CPU@3.1 GHz and 4 GB RAM.

*B. Experiment Results*

*1) Results on AUSLAN2 Dataset:* First, the effectiveness of using the RRV descriptor for single rigid body motion recognition will be evaluated on AUSLAN2 dataset so as to compare with the existing ISA and SoSaLe descriptors. The mean and the standard derivation (STD) CS recognition accuracies are recorded in Table III.

As can be observed, the results provided by our RRV de-scriptors (RRV-$l_2$+DTW, RRV+BoW, and RRV-new+DTW) outperform those with ISA (82.84±2.64%) and SoSaLe (73.27±2.43%) descriptors. The improvement mainly attributes to the descriptive power of the gradient-based RRV descriptor. Moreover, compared to $l_2$-norm, the proposed flexible metric also improves the recognition performance by computing the translational and rotational invariants separately. As demonstrated, the highest recognition rate achieves 92.56±2.04% with the setup RRV-new+DTW. Without considering temporal information, the result by RRV+BoW (90.92±2.61%) is slightly inferior to the highest accuracy.

Table IV reports the computational cost provided by different approaches. By choosing RRV+BoW, the time cost for recognizing each testing sample consumes 24 ms, which emphasizes that the BoW-based approach can balance the tradeoff between the recognition accuracy and the computational cost, which is beneficial for real-time applications.

Recall that the construction of the RRV descriptor only involves first-order time derivatives of the motion trajectory, which is insensitive to noise than the previous descriptors. To evaluate this, all 6-D rigid body motion trajectories will be added by white Gaussian noise with signal noise ratio (SNR) from 10 to 50. Fig. 10 shows the comparison results that contain the mean recognition accuracies provided by RRV-new, ISA, and SoSaLe descriptors in terms of the DTW-based recognition method. The blue square and black circle indicate the mean result by ISA and SoSaLe descriptors, and they both drastically decrease as SNR decreases. On the contrary, the results achieved by the RRV-new (red triangle) show strong resistance to noise.

*2) Results on MSRAction3D Dataset:* We then evaluate the extensions of the RRV descriptor for characterizing motions

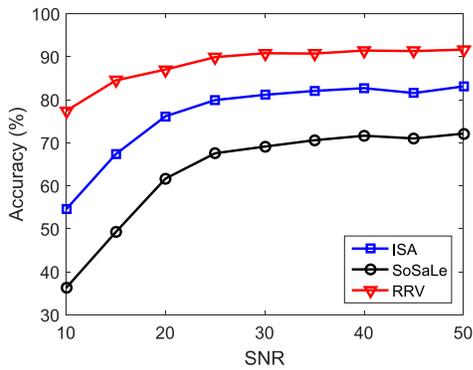

Fig. 10. Mean recognition accuracies on AUSLAN2 dataset while using different descriptors for representing noisy trajectories.

TABLE V
COMPARISON RESULTS WITH EXISTING SKELETON-BASED
METHODS ON MSRACTION3D DATASET

| Methods | Average(%) |
|---|---|
| Actionlet [22] | 82.33 |
| Lie group [8] | 89.48 |
| Pose set [23] | 90.22 |
| Cov3DJ [7] | 90.53 |
| Moving pose [24] | 91.70 |
| Multi-kernel learning [37] | 92.30 |
| RRV+LCS+DTW | 87.18 |
| RRV-VRB+GCS+DTW | 91.57 |
| RRV-VRB+LCS+BoW | 91.21 |
| **RRV-VRB+LCS+DTW** | **93.44** |

of a skeleton modeled as articulated interconnections of multiple rigid bodies on MSRAction3D dataset, and compare our approach with other state-of-the-art 3-D skeleton-based recognition works.

Table V reports the recognition results by various skeleton-based approaches on the MSRAction3D dataset. As can be observed, our approaches achieve better performance than the previous works. The highest recognition accuracy is 93.44% achieved by RRV-VRB+LCS+DTW, which outperforms the second best (92.30%) by 1.14%. Compared with the result using RRV+LCS+DTW, the discriminative VRB configuration can significantly improve the recognition performance. We also demonstrate the results by the setups RRV-VRB+GCS+DTW and RRV-VRB+LCS+DTW, and this comparison illustrates that the human body coordinate system $\{H\}$ can provide a convenient method to represent the joint trajectories in a view-independent local coordinate system and enhance the descriptiveness power for action recognition. This is because that the relations between different parts are not taken into consideration in RRV-VRB+GCS+DTW, thus leading to the result that similar actions cannot be well classified. Without considering temporal information, the result provided by RRV-VRB+LCS+BoW (91.21%) is inferior to that using DTW-based recognition approach.

Fig. 11(a) is the confusion matrix when recognition rate equals to 93.44%. As can be observed, most of the actions can be well classified, whereas the performance on some similar actions (*hammer*, *hand catch*, and *high throw*) is still needed to improve. The recent research [19] has reported the recognition accuracies as 100% on MSRAction3D dataset by using depth images. It should be noted that this paper focuses on exploring a novel skeleton representation modeled as the interconnections of multiple rigid bodies and incorporates our RRV descriptor for skeleton-based human action recognition. The time cost for recognizing each testing action sample consumes 607 ms with the DTW-based approach and 13 ms in terms of the BoW-based method.

TABLE VI
CROSS-VALIDATION RESULTS OF THE PROPOSED APPROACH
ON THE MSRACTION3D DATASET

| Methods | Mean±STD(%) |
|---|---|
| RRV-VRB+BoW | 85.23±2.92 |
| **RRV-VRB+LCS+DTW** | **87.76±3.58** |

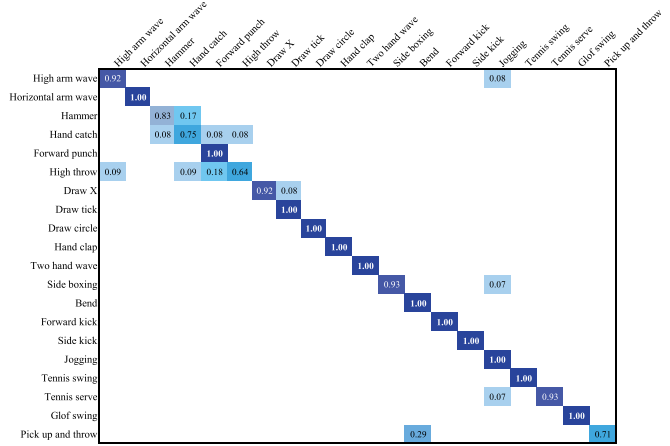

(a)

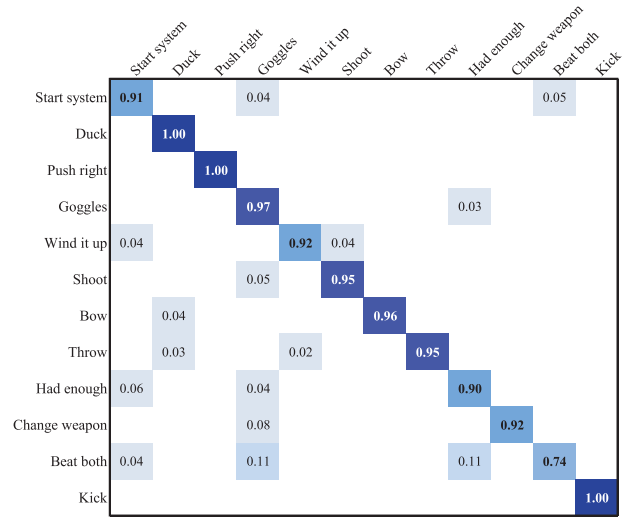

(b)

Fig. 11. Confusion matrices of the proposed approach regarding the highest recognition rates. (a) MSRAction3D dataset. (b) MSRC-12 dataset.

To demonstrate the consistency of our approach, Table VI reports the mean and STD results under the more challenging cross-validation protocol, where most of the previous works did not present this result. As can be seen, it can achieve 87.76±3.58% by RRV-VRB+LCS+DTW.

*3) Results on MSRC-12 Dataset:* Table VII gives the classification rates on MSRC-12 dataset using different

TABLE VII
COMPARISON RESULTS WITH EXISTING SKELETON-BASED
METHODS ON MSRC-12 DATASET

| Methods | Average(%) LOSubO | Cross-subject |
|---|---|---|
| ConvNets [25] | - | 93.12 |
| Cov3DJ [7] | 93.60 | 93.17 |
| RDF-selected features [42] | 94.03 | - |
| RRV-VRB+BoW | 92.34 | 92.89 |
| **RRV-VRB+LCS+DTW** | **94.71** | **93.87** |

skeletal-based human action representations. As can be seen, the setup RRV-VRB+LCS+DTW achieves 94.71% with LOSubO protocol, which outperforms the second best result (94.03% using RDF selected features [42]) by 0.68%. Due to the loss of temporal information, the recognition rate by RRV-VRB+BoW (92.34%) is slightly inferior to RRV-VRB+LCS+DTW.

For more challenging CS protocol, the best result is also achieved by RRV-VRB+LCS+DTW, where the recognition accuracy is 93.87%. In addition, the result by RRV-VRB+BoW is 92.89%. Although the recognition performance is inferior without considering the temporal information of the RRV descriptor sequences, the BoW-based recognition method only consumes about 20 ms for recognizing each testing sample, which is beneficial for real-time applications. Fig. 11(b) shows the confusion matrix when the recognition rate equals to 93.87% under CS protocol.

## VII. CONCLUSION

This paper devises an RRV descriptor for representing 6-D rigid body motion trajectories by exploring the local translational and rotational invariants. Our descriptor is invariant to rigid transformation and scale. Compared to the previous descriptors, the main contribution of our RRV descriptor is that it only involves first-order time derivatives of discrete trajectories, which is simple to construct and insensitive to noise. Instead of the $l_2$-norm, a flexible metric is also introduced to measure the distance between two RRV descriptors. Another contribution of this paper is that we extend the proposed RRV descriptor to the applications of multiple rigid bodies in skeleton-based human action recognition. A skeleton can be intuitively decomposed into five hierarchical body parts. The VRB is proposed for representing multiple rigid bodies in each part, which is discriminative and compact in the description. For rigid body motion recognition, we present a DTW-based recognition approach and a computational efficient BoW-based recognition approach. Finally, the experiments on AUSLAN2, MSRAction3D, and MSRC-12 datasets emphasize the effectiveness of our RRV descriptor. In the conclusion, our RRV descriptor can be applied in solving various rigid body motion recognition tasks.

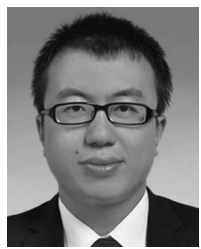

**Yao Guo** received the B.S. and M.S. degrees from Sun Yat-sen University, Guangzhou, China, in 2011 and 2014, respectively. He is currently pursuing the Ph.D. degree with the Department of Mechanical and Biomedical Engineering, City University of Hong Kong, Hong Kong, under the supervision of Prof. Y. Li.

His current research interests include pattern recognition, machine learning, robot sensing, and robot vision.

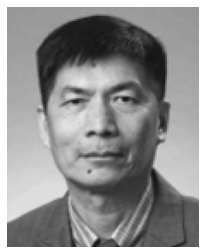

**Youfu Li** (M'91–SM'01) received the B.S. and M.S. degrees in electrical engineering from the Harbin Institute of Technology, Harbin, China, and the Ph.D. degree in robotics from the Department of Engineering Science, University of Oxford, Oxford, U.K., in 1993.

From 1993 to 1995, he was a Research Staff with the Department of Computer Science, University of Wales, Aberystwyth, U.K. He joined the City University of Hong Kong, Hong Kong, in 1995, where he is currently a Professor with the Department of Mechanical and Biomedical Engineering. His current research interests include robot sensing, robot vision, 3-D vision, and visual tracking.

Dr. Li has served as an Associate Editor for the IEEE TRANSACTIONS ON AUTOMATION SCIENCE AND ENGINEERING. He is currently serving as an Associate Editor for the IEEE ROBOTICS AND AUTOMATION MAGAZINE. He is an Editor of the IEEE Robotics and Automation Society Conference Editorial Board, the IEEE Conference on Robotics and Automation.

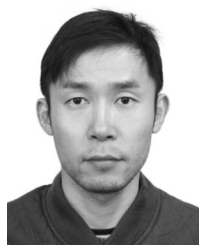

**Zhanpeng Shao** received the B.S. and M.S. degrees in mechanical engineering from the Xi'an University of Technology, Xi'an, China, in 2004 and 2007, respectively, and the Ph.D. degree in computer vision from the City University of Hong Kong, Hong Kong, in 2015.

He joined the Zhejiang University of Technology, Hangzhou, China, in 2016, where he is currently an Associate Professor with the Department of Computer Science and Technology. His current research interests include computer vision, pattern recognition, machine learning, and robot sensing.